\PassOptionsToPackage{table}{xcolor}
\documentclass[sigconf]{acmart}

\usepackage{algorithm}
\usepackage{algorithmic}
\usepackage{multirow}
\usepackage{caption}
\usepackage{tikz}
\definecolor{rankfirst}{HTML}{FFCCCC}
\definecolor{ranksecond}{HTML}{CCE5FF}
\definecolor{proxydark}{HTML}{333333}
\definecolor{degfull}{HTML}{F7FBFF}
\definecolor{degvgood}{HTML}{DEEBF7}
\definecolor{deggood}{HTML}{C6DBEF}
\definecolor{degfair}{HTML}{9ECAE1}
\definecolor{degpoor}{HTML}{6BAED6}
\definecolor{degbad}{HTML}{3182BD}
\definecolor{degworst}{HTML}{08519C}
\AtBeginDocument{%
  }

\setcopyright{none}
\renewcommand\footnotetextcopyrightpermission[1]{}

\begin{document}

\title{CoEvoP\&R: Co-Evolving Placement Objectives with Routing Feedback via Large Language Models}

\author{Ruogu Chen}
\orcid{0009-0009-1114-824X}
\affiliation{%
  \institution{University of Alberta}
  \city{Edmonton}
  \state{Alberta}
  \country{Canada}}
\email{ruogu@ualberta.ca}

\author{Weihua Xiao}
\authornote{Corresponding author.}
\orcid{0000-0002-6232-4460}
\affiliation{%
  \institution{New York University}
  \city{New York}
  \state{New York}
  \country{United States}}
\email{wx2356@nyu.edu}

\author{Ramesh Karri}
\orcid{0000-0001-7989-5617}
\affiliation{%
  \institution{New York University}
  \city{New York}
  \state{New York}
  \country{United States}}
\email{rkarri@nyu.edu}

\author{Jie Han}
\orcid{0000-0002-8849-4994}
\affiliation{%
  \institution{University of Alberta}
  \city{Edmonton}
  \state{Alberta}
  \country{Canada}}
\email{jhan8@ualberta.ca}

\renewcommand{\shortauthors}{Chen et al.}

\begin{abstract}

Analytical placers rely on differentiable objective functions to guide placement,
typically combining intermediate surrogate metrics such as
\textit{half-perimeter wirelength} (\textit{HPWL}) and cell-density penalties.
However, these placement-stage surrogates remain misaligned with downstream
routed and timing quality. Prior work reduces this gap with human-designed terms
or learned black-box surrogates, but the former requires expert retuning and the
latter is difficult to explain, debug, or deploy in analytical placement flows.
CoEvoP\&R addresses these limitations with a \textit{large language model}
(\textit{LLM})-based framework that automatically evolves analytical placement
objectives. At each generation, the prompt combines the restricted objective
interface, baseline context, and archived prior candidates with routing-related feedback 
from placement, timing proxy, and routing tools. The LLM
proposes readable differentiable objectives, which are embedded and validated
in DREAMPlace, evaluated through a timing proxy and an actual router, and stored with
their feedback to guide later generations.
Across eight ChiPBench Nangate45 designs and three seeds, CoEvoP\&R reduces
post-route routed wirelength and congestion by 16.9\%
and 36.7\%, with gains of
0.70 ns in worst negative slack and a 912 ns reduction
in total negative slack magnitude over native
DREAMPlace. Across
eight ICCAD 2015 Superblue designs, it reduces post-route
routed wirelength and congestion by 5.4\% and 23.2\%. Code is available at \url{https://github.com/FCHXWH823/CoEvoP-R.git}.

\end{abstract}




\maketitle
\pagestyle{plain}

\section{Introduction}

Modern analytical placers make global placement tractable by minimizing a
differentiable placement objective. Existing objectives usually combine
intermediate surrogates, such as smooth approximations to
\textit{half-perimeter wirelength} (\textit{HPWL}) and cell-density penalties,
which provide differentiable guidance for tools such as
DREAMPlace~\cite{dreamplace}. However, ChiPBench
shows that placement-stage metrics can remain weakly aligned with downstream
timing, including \textit{worst negative slack} (\textit{WNS}) and
\textit{total negative slack} (\textit{TNS})~\cite{chipbench}. Consequently,
optimizing placement surrogates does not necessarily optimize the final
\textit{power, performance, and area} (PPA) metrics of the completed
physical-design flow.

Prior work narrows this gap by combining the placement objective with
human-designed terms or learned black-box surrogates for the routing stage.
Human-designed terms, such as \textit{Rectangular Uniform wire DensitY}
(\textit{RUDY}), estimate
routing demand from placement geometry~\cite{rudy}, and subsequent methods
incorporate related congestion and pin-density information. These terms improve
routing awareness, but they rely on expert-designed approximations and often
require retuning across designs or technologies. Learned predictors,
post-route-informed density targets, and differentiable routability surrogates,
as used by LaMPlace~\cite{lamplace}, GOALPlace~\cite{goalplace}, and
RoutePlacer~\cite{routeplacer}, respectively, improve cross-stage awareness.
However,
black-box surrogates can be difficult to explain, debug, and deploy inside
existing analytical placement flows. This leaves a complementary path that
searches directly over readable, differentiable objective forms.

\textit{Large language models} (\textit{LLMs}) have shown potential in
directing the search for programs that satisfy or optimize specified targets.
FunSearch~\cite{funsearch}, Eureka~\cite{eureka}, LLM-SR~\cite{llmsr}, and
AlphaEvolve~\cite{alphaevolve} realize this idea by using evaluator feedback to
revise executable programs toward measured objectives.
EvoPlace~\cite{evoplace} and VeoPlace~\cite{veoplace} further apply foundation
models to placement by directing the search over optimizer components or
macro-placement actions.

Instead, we propose CoEvoP\&R, an LLM-based framework that automatically
evolves the analytical placement objective. For each generation, CoEvoP\&R
assembles a prompt from the restricted objective interface, baseline placement
context, archive records of earlier objective variants, and measured feedback
from placement-stage, timing-signal, and routed evaluations. The prompt guides the LLM
to generate the next candidate objectives.
CoEvoP\&R embeds and validates each admitted candidate in the analytical
placer, schedules downstream routed evaluation for selected candidates, and
stores the paired objective and measured feedback in the archive for subsequent
generations.
CoEvoP\&R makes four novel contributions.
\begin{itemize}
  \item \textbf{LLM objective evolution.} We present the
  first automated LLM-based framework that evolves analytical placement
  objective functions integrated with real placement and routing tools.

  \item \textbf{Automated validation.} CoEvoP\&R
  embeds and validates admitted objective programs inside DREAMPlace through a
  restricted interface, ensuring that generated objectives remain differentiable,
  bounded, and executable in an analytical placement flow.

  \item \textbf{Routing-aware feedback.} CoEvoP\&R evaluates
  evolved candidates with routing-tool feedback, including timing-related
  behavior, so the objective search is guided by downstream timing-related
  information.

  \item \textbf{Archive-guided iteration.} CoEvoP\&R stores each
  evolved candidate together with its placement, routing, timing, and failure
  feedback, then uses this archive to condition later prompts and guide
  subsequent generations.
\end{itemize}
Across eight ChiPBench Nangate45 designs, CoEvoP\&R reduces
routed wirelength and congestion by 16.9\%
and 36.7\% over native DREAMPlace, while improving post-route WNS by 0.70 ns
and TNS by 912 ns. Across eight Superblue designs, CoEvoP\&R reduce rWL and Cong.\ by 5.4\% and 23.2\% over the same baseline.

\section{Related Work}

\paragraph{Analytical and Routability-Aware Placement.}
ePlace developed a nonlinear placement framework based on electrostatics~\cite{eplace}.
RePlAce advanced its optimization and routability validation, and DREAMPlace
accelerated analytical placement with GPU-based PyTorch execution~\cite{replace,dreamplace}.
Timing-driven placers add timing guidance while retaining predefined objective
structures~\cite{dreamplace4,openroad}. Routability-aware placers incorporate
downstream information through routing estimators or learned guidance. RUDY
estimates routing demand from placement geometry~\cite{rudy}. LaMPlace learns
cross-stage metric masks for sequential macro placement, GOALPlace derives
density targets from post-route feedback, and RoutePlacer integrates a
differentiable graph neural network routability surrogate into analytical
placement~\cite{lamplace,goalplace,routeplacer}. These methods improve
downstream awareness through estimators, masks, targets, or surrogates, while
leaving the complete symbolic objective form outside the search.

\paragraph{Automated Placement Search.}
Automated placement methods differ in the artifact they search.
Reinforcement-learning methods learn sequential macro actions, AutoDMP and
HyperPlace tune placer parameters, and VeoPlace uses a vision-language model
to guide evolutionary macro placement~\cite{mirhoseini2021,autodmp,hyperplace,veoplace}.
EvoPlace is the closest placement-specific example of LLM-guided program
evolution. It searches initialization, preconditioning, and optimizer-update
components under HPWL feedback while retaining the placement
objective~\cite{evoplace}. These works show that automated search can improve actions, parameters, and optimizer behavior, leaving differentiable objective-form search under
downstream physical-design feedback as a distinct setting.

\paragraph{Executable Program Evolution.}
FunSearch and Evolution of Heuristics evolve executable heuristics, LLM-SR
searches symbolic equations, and AlphaEvolve extends evaluator-guided evolution
to broader code artifacts~\cite{funsearch,eoh,llmsr,alphaevolve}. OpenEvolve
provides an open implementation of this pattern~\cite{openevolve}. Eureka is
methodologically close because generated reward functions execute inside a
learning loop and later proposals use training feedback~\cite{eureka}.
Differentiable placement introduces additional requirements because accepted
expressions must produce finite values and gradients, adapt to placement trajectories, and justify expensive downstream evaluation.

\begin{figure*}[!t]
\centering
\includegraphics[width=\textwidth]{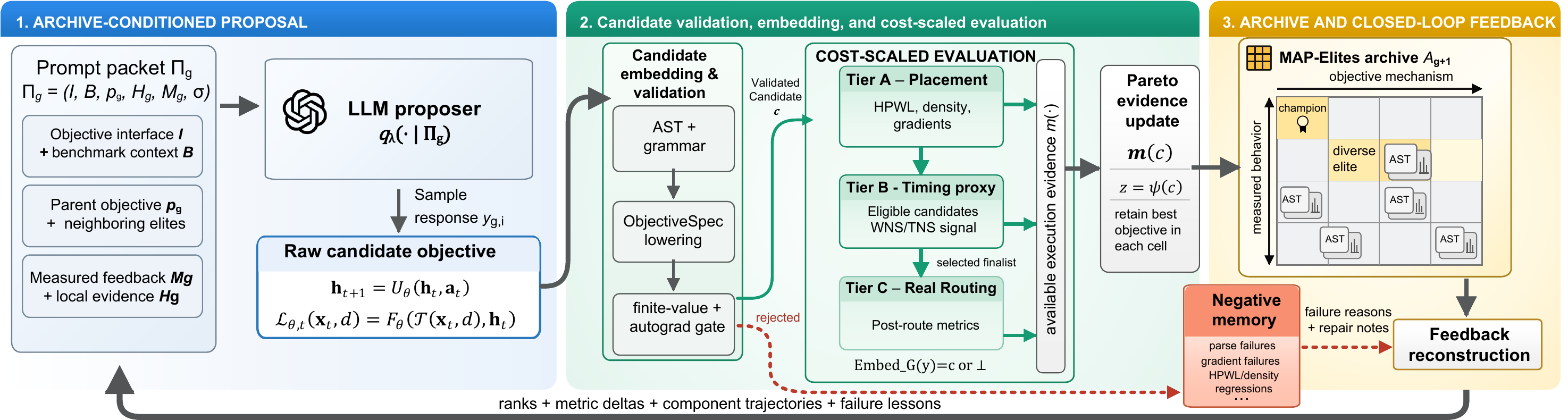}
\caption{CoEvoP\&R evolves complete placement objectives through
archive-conditioned proposal, DREAMPlace embedding, cost-scaled evaluation,
and archive feedback.}
\Description{A large language model proposes objective programs, validation embeds admitted
candidates into DREAMPlace, cost-scaled evaluations measure placement,
timing-signal, and routed evidence, and an archive reconstructs feedback for
later proposals.}
\label{fig-loop}
\end{figure*}

\section{Preliminaries}
\label{sec-preliminaries}

\subsection{Analytical Placers and Cross-Stage Metrics}
\label{sec-prelim-placement-metrics}

Chip placement assigns physical locations to circuit objects inside a fixed die.
A netlist can be modeled as a hypergraph $G=(V,E)$, with vertices $V$
representing macros, standard cells, and ports, and hyperedges $E$
representing nets. A placement is the coordinate
set $\mathbf{x}=\{(x_i,y_i)\}_{i\in V}$ after applying object sizes and fixed
pin offsets. Analytical placers optimize these coordinates with differentiable
objectives so that gradient-based updates can move many objects jointly.

DREAMPlace performs global placement with a smooth wirelength and
electrostatic-density objective~\cite{eplace,dreamplace}. At placement step
$t$, its native objective can be written as
\begin{equation}
\label{eq-prelim-native-placement}
\mathcal{L}_{\mathrm{DP},t}(\mathbf{x})
=
\widetilde{W}_{\gamma_t}(\mathbf{x})
+
\lambda_t D(\mathbf{x}) ,
\end{equation}
where $\widetilde{W}_{\gamma_t}$ is a differentiable wirelength approximation,
$D$ is the electrostatic density penalty, $\gamma_t$ controls wirelength
smoothing, and $\lambda_t$ balances wire shortening against density spreading.
The optimizer updates cell coordinates using the current gradient
$\nabla_{\mathbf{x}}\mathcal{L}_{\mathrm{DP},t}$. Across placement, the native
DREAMPlace objective forms a scheduled objective trajectory through the schedule variables $\gamma_t$ and $\lambda_t$.

After placement, HPWL is commonly used to evaluate the placement quality by: 
\begin{equation}
\label{eq-prelim-hpwl}
\operatorname{HPWL}(\mathbf{x})
=\sum_{e\in E}
\left[
\max_{i\in e} x_i-\min_{i\in e} x_i
+\max_{i\in e} y_i-\min_{i\in e} y_i
\right].
\end{equation}
HPWL decomposes over nets, tracks the bounding box of each net, and is fast to compute. Accordingly, it is widely used as a placement-stage metric.

However, placement-stage metrics are useful but incomplete~\cite{chipbench}. The routing demand, pin access, timing paths, buffering, and layer assignment are measured more
faithfully by downstream evaluations. Global routing reports routed wirelength and
routing overflow. Timing analysis reports WNS and TNS. These cross-stage metrics are closer to final design quality but require much more expensive physical design flows. During analytical placement, the optimizer instead operates on differentiable observables that are available throughout the placement trajectory.

\subsection{Differentiable Placement Observables}
\label{sec-prelim-observables}

A placement observable is a differentiable signal computed from the current
placement. Analytical placers partition the layout region into bins
$\mathcal{B}$, and $b\in\mathcal{B}$ indexes one bin. An observable can be
either a scalar or a bin map,
\begin{equation}
\label{eq-prelim-observable}
\phi(\mathbf{x})\in\mathbb{R}, \qquad
\Phi(\mathbf{x})=\{\Phi_b(\mathbf{x})\}_{b\in\mathcal{B}} .
\end{equation}
The gradients $\nabla_{\mathbf{x}}\phi$ and $\nabla_{\mathbf{x}}\Phi_b$ are
available during placement, allowing the optimizer to update cell coordinates
through these derivatives.

Scalar observables include smooth wirelength $\widetilde{W}(\mathbf{x})$ and
density penalties $D(\mathbf{x})$. Map observables include density maps
$\Phi^{\mathrm{den}}(\mathbf{x})$, RUDY-like routing demand maps
$\Phi^{\mathrm{route}}(\mathbf{x})$~\cite{rudy}, pin-density maps
$\Phi^{\mathrm{pin}}(\mathbf{x})$, and long-net pressure maps
$\Phi^{\mathrm{long}}(\mathbf{x})$.
These maps approximate physical pressure from placement geometry before routing
is performed.

Map observables are reduced to scalar terms before entering an analytical
objective. For a nonnegative map
$M(\mathbf{x})=\{M_b(\mathbf{x})\}_{b\in\mathcal{B}}$, a common smooth
reduction is
\begin{equation}
\label{eq-prelim-map-reduction}
\rho_p(M)=
\left(
\frac{1}{|\mathcal{B}|}
\sum_{b\in\mathcal{B}}
\left(M_b+\epsilon\right)^p
\right)^{1/p},
\end{equation}
where $p>1$ controls hotspot sensitivity and $\epsilon>0$ keeps the expression
well defined. Larger $p$ values place more weight on high-pressure bins, while
smaller values behave closer to an average.

A differentiable placement objective can therefore combine scalar observables
and reduced map observables as
\begin{equation}
\label{eq-prelim-observable-objective}
\mathcal{L}(\mathbf{x})
=F\left(
\phi_1(\mathbf{x}),\ldots,\phi_m(\mathbf{x}),
\rho_{p_1}(\Phi_1),\ldots,\rho_{p_n}(\Phi_n)
\right),
\end{equation}
where $F$ is a differentiable composition of these terms.

\section{Method}
\label{sec-method}

CoEvoP\&R searches for readable differentiable placement objectives that run
inside an analytical placer while the placer, benchmarks, and downstream
evaluation flow remain fixed. Each candidate $\theta$ is a compact symbolic
objective that generalizes the static composition of
Equation~\ref{eq-prelim-observable-objective} into a scheduled one,
\begin{equation}
\label{eq-method-scheduled-form}
\mathbf{h}_0=I_{\theta}(\mathbf{a}_0),
\quad
\mathbf{h}_{t+1}=U_{\theta}(\mathbf{h}_t,\mathbf{a}_t),
\quad
\mathcal{L}_{\theta,t}(\mathbf{x}_t,d)
=F_{\theta}\!\left(\mathcal{T}(\mathbf{x}_t,d),\mathbf{h}_t\right).
\end{equation}
Here $\mathcal{T}(\mathbf{x}_t,d)$ collects differentiable placement terms,
$\mathbf{h}_t$ carries the time-varying coefficients we call schedule
variables, and $\mathbf{a}_t$ carries simple progress signals such as the
current density overflow and the recent wirelength trend. A candidate is
therefore fully specified by $I_{\theta}$, $U_{\theta}$, and $F_{\theta}$.
Across generations, evaluated candidates accumulate in an archive that
supplies parents and local evidence for later proposals.
Figure~\ref{fig-loop} organizes this closed loop into three steps:
archive-conditioned objective proposal, candidate embedding with cost-scaled
evaluation, and archive feedback for the next generation.

\begin{center}
\begingroup
\scriptsize
\setlength{\tabcolsep}{2.2pt}
\renewcommand{\arraystretch}{0.86}
\begin{tabular}{@{}p{0.11\columnwidth}p{0.35\columnwidth}p{0.11\columnwidth}p{0.35\columnwidth}@{}}
\toprule
\multicolumn{4}{@{}l@{}}{\textbf{Key method notation}} \\
\midrule
Symbol & Meaning & Symbol & Meaning \\
\midrule
$g$ & evolution generation & $t$ & placement iteration \\
$d$ & design being optimized & $\mathbf{x}_t$ & placement coordinates at iteration $t$ \\
$\mathcal{I}$ & objective interface & $B$ & baseline context \\
$\sigma$ & output schema & $\mathcal{A}_g$ & MAP-Elites archive at generation $g$ \\
$p_g$ & parent objective & $H_g$ & local archive evidence \\
$M_g$ & measured feedback & $\Pi_g$ & prompt packet \\
$y$ & proposed program & $c$ & validated candidate \\
$\bot$ & failed proposal & $\theta$ & candidate objective \\
$\mathbf{h}_t$ & schedule variables & $\mathbf{a}_t$ & progress observables \\ % a_t NEW
$I_\theta,U_\theta,F_\theta$ & init, update, and loss maps of $\theta$ &
$\mathcal{T}(\mathbf{x}_t,d)$ & differentiable placement terms \\ % maps NEW
$\mathbf{m}(c)$ & measured evidence for $c$ & $\psi(c)$ & archive feature mapping \\
$z$ & MAP-Elites cell & $\preceq$ & Pareto evidence order \\
$K$ & samples per generation & $\mathcal{N}$ & negative memory \\ % NEW row
\bottomrule
\end{tabular}
\endgroup
\end{center}

\subsection{Archive-Conditioned Objective Proposal}
\label{sec-method-objective-form}

\begin{figure}[t]
  \centering
  \includegraphics[width=\columnwidth]{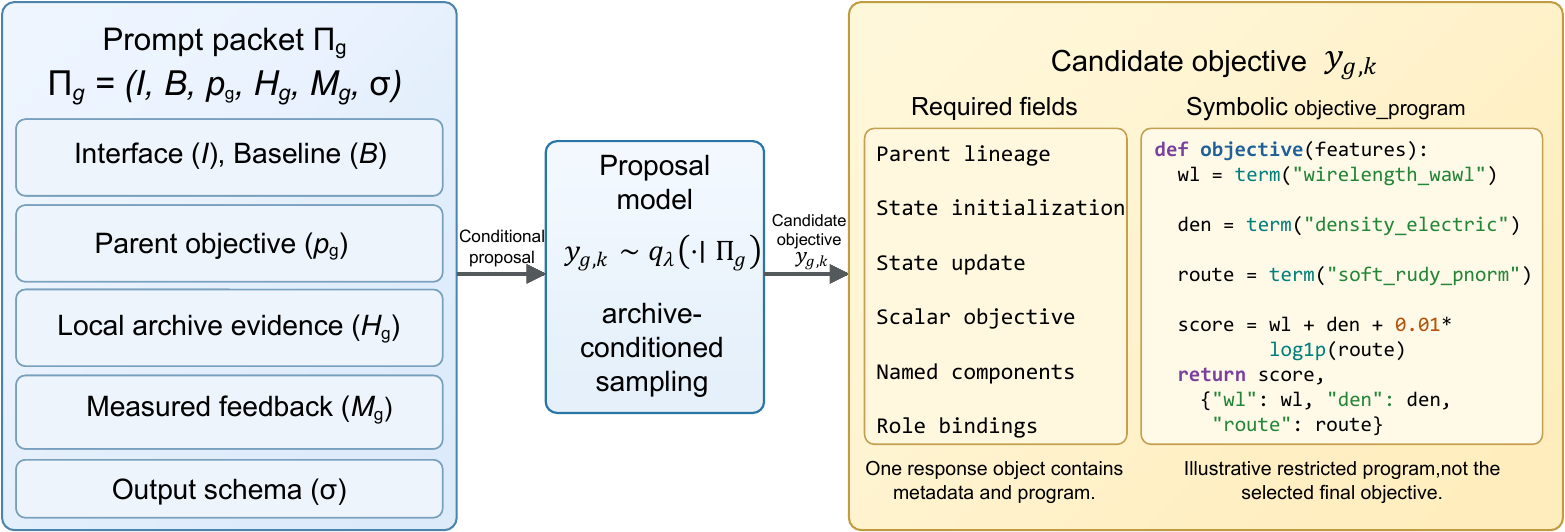}
  \caption{Archive-conditioned proposal contract.}
  \Description{The figure shows prompt packet construction, archive-conditioned
  proposal sampling, and the required fields of a sampled objective program.}
  \label{fig-proposal-embedding}
\end{figure}

At generation $g$, the current archive $\mathcal{A}_g$ supplies a parent
objective $p_g$ and a local evidence set $H_g$. The evidence set contains
neighboring elites, component trajectories, metric deltas, and failure lessons
from earlier runs. CoEvoP\&R pairs this evidence with measured feedback $M_g$
and the objective interface $\mathcal{I}$ to form the prompt packet
$\Pi_g=(\mathcal{I},B,p_g,H_g,M_g,\sigma)$, where $B$ gives baseline descriptions and
$\sigma$ gives the required output format. Figure~\ref{fig-proposal-embedding}
shows this proposal contract. The prompt packet conditions the proposal model,
and each response must provide the fields needed to instantiate a
trajectory-aware objective. Archive conditioning guides proposals toward
objective mechanisms that occupy useful behavioral cells and toward
measurements that explain nearby placement improvements or degradations.

Each response must therefore provide the three functions of
Equation~\ref{eq-method-scheduled-form}: an initialization $I_{\theta}$ that
sets the schedule variables from design-specific calibration, an update
$U_{\theta}$ that revises them once per placement iteration, and a loss
$F_{\theta}$ over the placement terms. The progress vector $\mathbf{a}_t$
exposes the iteration fraction, density overflow, HPWL trend, smoothing
position, and per-term calibration constants, so proposed schedules react to
measured optimization progress rather than to a fixed clock.

The runtime objective embedded in DREAMPlace has the form
\begin{equation}
\label{eq-method-runtime-loss}
\begin{aligned}
\mathcal{L}_{\theta,t}
&=\widetilde{W}_{\gamma_t}(\mathbf{x}_t)+h^{\mathrm{den}}_tD(\mathbf{x}_t)
+h^{\mathrm{route}}_tR(\mathbf{x}_t)+h^{\mathrm{pin}}_tP(\mathbf{x}_t),\\
\gamma_t&=h^{\gamma}_t\gamma^{\mathrm{base}}_t .
\end{aligned}
\end{equation}
The term $\widetilde{W}_{\gamma_t}$ is the smoothed wirelength used by the
placer, $D$ is the density penalty, $R$ is a differentiable routing pressure
term, and $P$ is a pin pressure term. The state variables
$h^{\mathrm{den}}_t$, $h^{\mathrm{route}}_t$, $h^{\mathrm{pin}}_t$, and
$h^\gamma_t$ determine the relative weights and the active smoothing scale.
This formulation makes archive evidence actionable through complete objectives
whose coefficients and smoothing behavior evolve with placement progress.

\subsection{Candidate Validation and DREAMPlace Embedding}
\label{sec-method-embedding}

The proposal model produces symbolic objectives, while DREAMPlace requires
terms, components, and state updates that are bound to placement quantities.
CoEvoP\&R performs this validation and embedding as the middle stage of
Figure~\ref{fig-loop}.

Let $\operatorname{Embed}_d$ denote the embedding procedure for design $d$.
For a proposed symbolic program $y$, this procedure either returns a candidate
$c$ that can be executed in DREAMPlace or marks the proposal as failed.
Formally,
\begin{equation}
\label{eq-method-embedding}
\operatorname{Embed}_d(y)=
\begin{cases}
c, & y\in\mathcal{Y}_d,\\
\bot, & y\in\overline{\mathcal{Y}}_d .
\end{cases}
\end{equation}
Here $\mathcal{Y}_d$ is the set of proposed programs accepted by the grammar,
physical-role, differentiability, and runtime checks for design $d$.
The complement $\overline{\mathcal{Y}}_d$ contains rejected programs, and $\bot$
denotes an invalid candidate that is not executed by DREAMPlace. A candidate
$c$ contains the initialization map, update map, scalar loss, named components,
and schedule variables required by Equations~\ref{eq-method-scheduled-form}
and~\ref{eq-method-runtime-loss}.

After embedding, candidate $c$ enters DREAMPlace through the physical roles in
Equation~\ref{eq-method-runtime-loss}. It provides the density, routing, pin,
and smoothing schedules, while native wirelength and density operators form
the differentiable placement backbone. The density state also updates the
placer preconditioner, so objective changes affect both the scalar loss and
the optimization trajectory. Timing-aware candidates may adjust net weights
using placement-time criticality, span, and fanout features.

% Validated candidates enter placement after CoEvoP\&R checks finite scalar
% values, available gradients, valid state updates, bounded smoothing,
% nonnegative physical weights, and completion of a DREAMPlace run.
% For every admitted objective, the placer records component traces, state
% trajectories, HPWL, overflow, and produced placement artifacts. These traces
% expose each objective as a physical optimization process with measurable
% intermediate behavior.

\subsection{Cost-Scaled Evaluation and Evidence Ranking}
\label{sec-method-evaluation}
\label{sec-proxy-audit}

Downstream routing and timing tools provide faithful feedback at substantially
higher cost than placement. CoEvoP\&R therefore evaluates candidates
through a cost-scaled evidence hierarchy. Tier A runs the analytical placer for every validated candidate and records
HPWL, overflow, runtime, convergence behavior, and objective traces. Tier B
applies a placement-stage timing proxy to selected candidates and records
proxy WNS/TNS movement. Tier C schedules the downstream router for
real routing and records routed wirelength, routing overflow, and post-route WNS
and TNS. The evidence vector for a validated candidate $c$ is
\[
\mathbf{m}(c)=m_A(c)\oplus a_B(c)m_B(c)\oplus a_C(c)m_C(c),
\]
where $a_B$ and $a_C$ indicate whether the candidate receives the corresponding
downstream measurement.

Archive selection uses Pareto evidence over the measurements available to each
candidate pair. For candidate $c$, define
\begin{equation}
\label{eq-method-pareto}
\begin{aligned}
\mathbf{v}(c)
&=\left(\Delta_{\mathrm{HPWL}}(c),
\Delta_{\mathrm{ovf}}(c),
-\Delta_{\mathrm{WNS}}(c),
-\Delta_{\mathrm{TNS}}(c)\right),\\
c_i\prec c_j
&\Longleftrightarrow
v_k(c_i)\le v_k(c_j)\ \forall k\in\mathcal{K}_{ij}\\
&\quad\land\
v_k(c_i)<v_k(c_j)\ \exists k\in\mathcal{K}_{ij}.
\end{aligned}
\end{equation}
The vector $\mathbf{v}(c)$ uses lower-is-better coordinates, with timing signs
chosen so that WNS and TNS improvements align with HPWL and overflow
improvements. The index set $\mathcal{K}_{ij}$ contains the coordinates
measured for both candidates. We write $c_i\preceq c_j$ when $c_i\prec c_j$
holds or no measured coordinate differs. Structural failures, invalid placements,
nonfinite traces, and robustness failures are removed before Pareto ranking.
Candidates on the same nondominated front are ordered by design coverage and
then by runtime.

The timing proxy audit controls how placement-stage timing evidence
influences the archive. CoEvoP\&R periodically applies controlled perturbations
to placement terms, net-weighting choices, and state schedules, then compares
the resulting placement-stage WNS/TNS movement with downstream WNS/TNS
movement using Pearson and Spearman redundancy checks. A timing proxy that
adds stable information beyond HPWL and overflow receives more influence in
archive updates. An unstable or redundant timing proxy remains available to
objective construction but receives little authority in evidence ranking.

\begin{table*}[!t]
\centering
\caption{Main ChiPBench Nangate45 results and intra-family transfer.
Entries report three-seed mean improvements over DREAMPlace, which is normalized to zero. rWL and
Cong.\ are reductions in \textit{routed wirelength} and post-route
\textit{congestion}. WNS and TNS are post-route timing gains in ns. 
\colorbox{rankfirst}{\textbf{Bold}} marks best means and
\colorbox{ranksecond}{\underline{underline}} marks second best.}
\label{tab:chipbench-main}
\scriptsize
\renewcommand{\arraystretch}{0.76}
\setlength{\tabcolsep}{3pt}
\begin{tabular}{ll*{9}{r}}
\toprule
& & \multicolumn{2}{c}{BlackParrot family}
& \multicolumn{3}{c}{Processor cores}
& \multicolumn{3}{c}{Peripheral and control}
& \multicolumn{1}{c}{Mean} \\
\cmidrule(lr){3-4}\cmidrule(lr){5-7}\cmidrule(lr){8-10}\cmidrule(l){11-11}
Method & Metric & bp\_fe & bp\_be & swerv\_wrapper & or1200 & mor1kx & ethernet & dft68 & vga\_lcd & Mean \\
\midrule
\multicolumn{11}{@{}l}{\emph{Analytical, expert, and search baselines}} \\
\multirow{4}{*}{DREAMPlace 4.0~\cite{dreamplace4}}
& rWL (\%) & -5.3$_{\pm0.3}$ & -1.7$_{\pm0.2}$ & -3.9$_{\pm0.3}$ & -4.6$_{\pm0.4}$ & -2.1$_{\pm0.2}$ & -0.4$_{\pm0.1}$ & -1.8$_{\pm0.2}$ & 0.3$_{\pm0.1}$ & -2.4$_{\pm0.2}$ \\
& Cong. (\%) & -2.4$_{\pm0.3}$ & 0.7$_{\pm0.1}$ & -1.8$_{\pm0.2}$ & -1.1$_{\pm0.2}$ & -1.2$_{\pm0.2}$ & -2.3$_{\pm0.3}$ & -0.4$_{\pm0.1}$ & 0.4$_{\pm0.1}$ & -1.0$_{\pm0.2}$ \\
& WNS (ns) & +0.38$_{\pm0.03}$ & +0.48$_{\pm0.04}$ & +0.18$_{\pm0.02}$ & \cellcolor{rankfirst}\textbf{+1.32}$_{\pm0.09}$ & +0.16$_{\pm0.02}$ & \cellcolor{rankfirst}\textbf{+0.36}$_{\pm0.03}$ & +0.12$_{\pm0.02}$ & +0.28$_{\pm0.03}$ & +0.41$_{\pm0.03}$ \\
& TNS (ns) & \cellcolor{ranksecond}\underline{+720}$_{\pm48}$ & +480$_{\pm32}$ & +140$_{\pm12}$ & +940$_{\pm62}$ & +220$_{\pm18}$ & +340$_{\pm24}$ & +70$_{\pm6}$ & +640$_{\pm42}$ & +444$_{\pm28}$ \\
\cmidrule(lr){1-11}
\multirow{4}{*}{Best expert}
& rWL (\%) & -2.4$_{\pm0.4}$ & 3.7$_{\pm0.3}$ & -0.8$_{\pm0.2}$ & 2.6$_{\pm0.4}$ & -2.3$_{\pm0.3}$ & 1.9$_{\pm0.2}$ & -1.2$_{\pm0.2}$ & 4.1$_{\pm0.4}$ & 0.7$_{\pm0.3}$ \\
& Cong. (\%) & 4.7$_{\pm0.4}$ & 6.8$_{\pm0.5}$ & 0.9$_{\pm0.2}$ & 4.2$_{\pm0.4}$ & 3.7$_{\pm0.3}$ & 3.4$_{\pm0.3}$ & 1.6$_{\pm0.2}$ & 2.7$_{\pm0.3}$ & 3.5$_{\pm0.3}$ \\
& WNS (ns) & -0.14$_{\pm0.03}$ & -0.21$_{\pm0.03}$ & -0.08$_{\pm0.02}$ & -0.28$_{\pm0.04}$ & -0.09$_{\pm0.02}$ & -0.18$_{\pm0.03}$ & -0.04$_{\pm0.01}$ & -0.03$_{\pm0.01}$ & -0.13$_{\pm0.02}$ \\
& TNS (ns) & -22.4$_{\pm2.6}$ & -34.6$_{\pm3.4}$ & -12.7$_{\pm1.8}$ & -178.6$_{\pm14.2}$ & -27.2$_{\pm3.2}$ & -68.4$_{\pm6.4}$ & -8.7$_{\pm1.4}$ & -13.4$_{\pm1.8}$ & -46$_{\pm4.2}$ \\
\cmidrule(lr){1-11}
\multirow{4}{*}{BO-DSL}
& rWL (\%) & -0.7$_{\pm0.8}$ & 3.8$_{\pm0.7}$ & 1.2$_{\pm0.6}$ & 2.9$_{\pm0.7}$ & -0.4$_{\pm0.4}$ & 4.6$_{\pm0.8}$ & -0.3$_{\pm0.4}$ & 5.7$_{\pm0.9}$ & 2.1$_{\pm0.6}$ \\
& Cong. (\%) & 6.3$_{\pm0.9}$ & 5.4$_{\pm0.7}$ & 2.8$_{\pm0.6}$ & 6.1$_{\pm0.8}$ & 5.4$_{\pm0.7}$ & 4.9$_{\pm0.6}$ & 2.7$_{\pm0.5}$ & 4.2$_{\pm0.6}$ & 4.7$_{\pm0.7}$ \\
& WNS (ns) & -0.06$_{\pm0.04}$ & +0.04$_{\pm0.02}$ & -0.09$_{\pm0.03}$ & -0.13$_{\pm0.05}$ & -0.01$_{\pm0.02}$ & -0.03$_{\pm0.02}$ & +0.01$_{\pm0.01}$ & +0.03$_{\pm0.02}$ & -0.03$_{\pm0.03}$ \\
& TNS (ns) & -12.4$_{\pm3.6}$ & +6.8$_{\pm2.4}$ & -18.7$_{\pm4.2}$ & -52.7$_{\pm7.8}$ & -6.4$_{\pm2.8}$ & -14.6$_{\pm3.4}$ & +4.2$_{\pm1.4}$ & +8.9$_{\pm2.2}$ & -10.6$_{\pm3.6}$ \\
\cmidrule(lr){1-11}
\multirow{4}{*}{AutoDMP~\cite{autodmp}}
& rWL (\%) & 7.8$_{\pm0.5}$ & 8.6$_{\pm0.4}$ & 11.3$_{\pm0.7}$ & 6.9$_{\pm0.4}$ & \cellcolor{rankfirst}\textbf{9.7}$_{\pm0.6}$ & 12.7$_{\pm0.8}$ & \cellcolor{ranksecond}\underline{5.4}$_{\pm0.3}$ & 14.8$_{\pm0.9}$ & 9.7$_{\pm0.6}$ \\
& Cong. (\%) & 9.4$_{\pm0.6}$ & 11.7$_{\pm0.7}$ & 7.8$_{\pm0.5}$ & 10.7$_{\pm0.6}$ & 12.9$_{\pm0.8}$ & 13.6$_{\pm0.8}$ & 8.4$_{\pm0.5}$ & 14.2$_{\pm0.9}$ & 11.1$_{\pm0.7}$ \\
& WNS (ns) & +0.27$_{\pm0.03}$ & +0.28$_{\pm0.03}$ & +0.08$_{\pm0.02}$ & +0.59$_{\pm0.04}$ & +0.14$_{\pm0.02}$ & \cellcolor{ranksecond}\underline{+0.34}$_{\pm0.03}$ & +0.05$_{\pm0.01}$ & +0.07$_{\pm0.02}$ & +0.23$_{\pm0.02}$ \\
& TNS (ns) & +112$_{\pm10}$ & +114$_{\pm12}$ & +30$_{\pm4}$ & +840$_{\pm62}$ & +80$_{\pm8}$ & +324$_{\pm28}$ & +14$_{\pm3}$ & +49$_{\pm6}$ & +195$_{\pm16}$ \\
\cmidrule(lr){1-11}
\multirow{4}{*}{EvoPlace~\cite{evoplace}}
& rWL (\%) & 2.4$_{\pm0.7}$ & 6.8$_{\pm0.6}$ & 5.1$_{\pm0.5}$ & 4.3$_{\pm0.5}$ & \cellcolor{ranksecond}\underline{9.5}$_{\pm0.7}$ & 9.7$_{\pm0.9}$ & 0.9$_{\pm0.3}$ & 14.6$_{\pm1.2}$ & 6.7$_{\pm0.7}$ \\
& Cong. (\%) & 1.8$_{\pm0.6}$ & 2.9$_{\pm0.4}$ & -0.6$_{\pm0.4}$ & 2.4$_{\pm0.5}$ & 0.7$_{\pm0.3}$ & 2.7$_{\pm0.5}$ & 0.3$_{\pm0.2}$ & 3.6$_{\pm0.6}$ & 1.7$_{\pm0.4}$ \\
& WNS (ns) & +0.04$_{\pm0.02}$ & +0.06$_{\pm0.02}$ & -0.03$_{\pm0.02}$ & +0.07$_{\pm0.02}$ & +0.02$_{\pm0.01}$ & +0.03$_{\pm0.02}$ & +0.01$_{\pm0.01}$ & -0.04$_{\pm0.02}$ & +0.02$_{\pm0.02}$ \\
& TNS (ns) & +8.4$_{\pm2.2}$ & +14.7$_{\pm3.2}$ & -4.2$_{\pm2.4}$ & +38.4$_{\pm5.6}$ & +6.3$_{\pm2.2}$ & +22.6$_{\pm3.8}$ & +2.8$_{\pm1.6}$ & -9.7$_{\pm2.8}$ & +9.9$_{\pm2.8}$ \\
\cmidrule(lr){1-11}
\multirow{4}{*}{LaMPlace~\cite{lamplace}}
& rWL (\%) & \cellcolor{rankfirst}\textbf{33.6}$_{\pm2.1}$ & \cellcolor{rankfirst}\textbf{18.9}$_{\pm1.4}$ & 8.9$_{\pm0.7}$ & 6.4$_{\pm0.5}$ & -2.4$_{\pm0.3}$ & -4.6$_{\pm0.5}$ & \cellcolor{rankfirst}\textbf{5.6}$_{\pm0.4}$ & -8.4$_{\pm0.7}$ & 7.3$_{\pm0.8}$ \\
& Cong. (\%) & \cellcolor{rankfirst}\textbf{33.1}$_{\pm2.1}$ & 20.6$_{\pm1.3}$ & 4.4$_{\pm0.3}$ & 24.6$_{\pm1.6}$ & 18.4$_{\pm1.2}$ & 22.8$_{\pm1.4}$ & 14.6$_{\pm0.9}$ & \cellcolor{rankfirst}\textbf{26.4}$_{\pm1.7}$ & 20.6$_{\pm1.3}$ \\
& WNS (ns) & +0.05$_{\pm0.01}$ & +0.19$_{\pm0.02}$ & +0.04$_{\pm0.01}$ & +0.22$_{\pm0.02}$ & +0.13$_{\pm0.02}$ & +0.28$_{\pm0.03}$ & +0.06$_{\pm0.01}$ & +0.24$_{\pm0.02}$ & +0.15$_{\pm0.02}$ \\
& TNS (ns) & -407$_{\pm28}$ & \cellcolor{rankfirst}\textbf{+1323}$_{\pm88}$ & +40$_{\pm4}$ & +680$_{\pm46}$ & \cellcolor{rankfirst}\textbf{+620}$_{\pm42}$ & +380$_{\pm26}$ & +48$_{\pm5}$ & +540$_{\pm36}$ & +403$_{\pm28}$ \\
\midrule
\multicolumn{11}{@{}l}{\emph{CoEvoP\&R evolved objectives}} \\
\multirow{4}{*}{\textbf{CoEvoP\&R-E}}
& rWL (\%) & \cellcolor{ranksecond}\underline{32.4}$_{\pm2.2}$ & \cellcolor{ranksecond}\underline{17.4}$_{\pm1.3}$ & \cellcolor{rankfirst}\textbf{18.4}$_{\pm1.3}$ & \cellcolor{rankfirst}\textbf{13.6}$_{\pm1.0}$ & 9.4$_{\pm0.7}$ & \cellcolor{rankfirst}\textbf{17.4}$_{\pm1.2}$ & 5.2$_{\pm0.4}$ & \cellcolor{rankfirst}\textbf{21.4}$_{\pm1.5}$ & \cellcolor{rankfirst}\textbf{16.9}$_{\pm1.0}$ \\
& Cong. (\%) & \cellcolor{ranksecond}\underline{31.4}$_{\pm2.0}$ & \cellcolor{rankfirst}\textbf{33.6}$_{\pm2.2}$ & \cellcolor{rankfirst}\textbf{68.4}$_{\pm3.8}$ & \cellcolor{rankfirst}\textbf{43.6}$_{\pm2.6}$ & \cellcolor{rankfirst}\textbf{28.4}$_{\pm1.8}$ & \cellcolor{rankfirst}\textbf{36.4}$_{\pm2.3}$ & \cellcolor{rankfirst}\textbf{27.2}$_{\pm1.7}$ & \cellcolor{ranksecond}\underline{24.6}$_{\pm1.6}$ & \cellcolor{rankfirst}\textbf{36.7}$_{\pm2.0}$ \\
& WNS (ns) & \cellcolor{rankfirst}\textbf{+0.62}$_{\pm0.05}$ & \cellcolor{rankfirst}\textbf{+0.88}$_{\pm0.07}$ & \cellcolor{rankfirst}\textbf{+0.42}$_{\pm0.04}$ & \cellcolor{ranksecond}\underline{+1.18}$_{\pm0.09}$ & \cellcolor{rankfirst}\textbf{+0.38}$_{\pm0.03}$ & +0.32$_{\pm0.03}$ & \cellcolor{rankfirst}\textbf{+0.72}$_{\pm0.06}$ & \cellcolor{rankfirst}\textbf{+1.07}$_{\pm0.09}$ & \cellcolor{rankfirst}\textbf{+0.70}$_{\pm0.06}$ \\
& TNS (ns) & \cellcolor{rankfirst}\textbf{+940}$_{\pm68}$ & \cellcolor{ranksecond}\underline{+1180}$_{\pm88}$ & \cellcolor{rankfirst}\textbf{+380}$_{\pm30}$ & \cellcolor{rankfirst}\textbf{+1560}$_{\pm118}$ & \cellcolor{ranksecond}\underline{+540}$_{\pm42}$ & \cellcolor{rankfirst}\textbf{+680}$_{\pm48}$ & \cellcolor{rankfirst}\textbf{+190}$_{\pm16}$ & \cellcolor{rankfirst}\textbf{+1830}$_{\pm136}$ & \cellcolor{rankfirst}\textbf{+912}$_{\pm68}$ \\
\cmidrule(lr){1-11}
\multirow{4}{*}{\textbf{CoEvoP\&R-L}}
& rWL (\%) & 25.4$_{\pm1.9}$ & 13.4$_{\pm1.2}$ & \cellcolor{ranksecond}\underline{14.4}$_{\pm1.1}$ & \cellcolor{ranksecond}\underline{10.4}$_{\pm0.8}$ & 8.4$_{\pm0.7}$ & \cellcolor{ranksecond}\underline{13.4}$_{\pm1.0}$ & 4.4$_{\pm0.3}$ & \cellcolor{ranksecond}\underline{15.8}$_{\pm1.2}$ & \cellcolor{ranksecond}\underline{13.2}$_{\pm0.9}$ \\
& Cong. (\%) & 28.4$_{\pm2.1}$ & \cellcolor{ranksecond}\underline{32.4}$_{\pm2.1}$ & \cellcolor{ranksecond}\underline{56.4}$_{\pm3.6}$ & \cellcolor{ranksecond}\underline{35.4}$_{\pm2.3}$ & \cellcolor{ranksecond}\underline{25.4}$_{\pm1.8}$ & \cellcolor{ranksecond}\underline{30.4}$_{\pm2.0}$ & \cellcolor{ranksecond}\underline{20.4}$_{\pm1.4}$ & 22.4$_{\pm1.6}$ & \cellcolor{ranksecond}\underline{31.4}$_{\pm1.9}$ \\
& WNS (ns) & \cellcolor{ranksecond}\underline{+0.46}$_{\pm0.05}$ & \cellcolor{ranksecond}\underline{+0.64}$_{\pm0.06}$ & \cellcolor{ranksecond}\underline{+0.32}$_{\pm0.04}$ & +1.02$_{\pm0.09}$ & \cellcolor{ranksecond}\underline{+0.21}$_{\pm0.03}$ & +0.24$_{\pm0.03}$ & \cellcolor{ranksecond}\underline{+0.48}$_{\pm0.04}$ & \cellcolor{ranksecond}\underline{+0.68}$_{\pm0.06}$ & \cellcolor{ranksecond}\underline{+0.51}$_{\pm0.04}$ \\
& TNS (ns) & +530$_{\pm42}$ & +1100$_{\pm86}$ & \cellcolor{ranksecond}\underline{+260}$_{\pm22}$ & \cellcolor{ranksecond}\underline{+1000}$_{\pm76}$ & +380$_{\pm32}$ & \cellcolor{ranksecond}\underline{+480}$_{\pm36}$ & \cellcolor{ranksecond}\underline{+128}$_{\pm12}$ & \cellcolor{ranksecond}\underline{+1220}$_{\pm92}$ & \cellcolor{ranksecond}\underline{+637}$_{\pm48}$ \\
\bottomrule
\end{tabular}
\end{table*}

\subsection{Archive Update and Closed-Loop Feedback}
\label{sec-method-archive}

The archive is organized as a MAP-Elites grid over objective
behavior~\cite{mapelites}. Each
admitted objective is mapped to a feature cell
\[
z=\psi(c),
\]
where $\psi$ summarizes complexity, mechanism family, HPWL behavior, overflow
behavior, and timing behavior. To preserve diversity, CoEvoP\&R uses an island
model in which archive subpopulations exchange elites periodically. Within
island $r$, the archive update is
\[
\mathcal{A}_{g+1}^{(r)}[z]=
\operatorname{elite}_{\preceq}\left(
\mathcal{A}_{g}^{(r)}[z]\cup\mathcal{C}_{g}^{(r)}[z]\right),
\]
where $\mathcal{C}_{g}^{(r)}[z]$ collects the admitted generation-$g$
candidates of island $r$ mapped to cell $z$, and $\preceq$ is the Pareto
evidence order from Equation~\ref{eq-method-pareto}. Negative memory stores failed candidates and
their failure reasons, so later prompt packets avoid recurring invalid
mechanisms. Island migration periodically exchanges elites across subpopulations
and exposes the proposal model to objectives that succeed under different
behavioral niches. 

Algorithm~\ref{alg-coevo-evolution} summarizes the loop that returns a readable objective with state variables, named components, lineage, and measured
placement evidence.

\begin{algorithm}[h]
\caption{CoEvo objective evolution}
\label{alg-coevo-evolution}
\small
\begin{algorithmic}[1]
\STATE $\mathcal{A}_0,\mathcal{N}\leftarrow
\operatorname{Init}(\text{seeds},\text{baselines}),\emptyset$
\FOR{$g=0,\ldots,G-1$}
\STATE $(r,p_g,H_g,M_g)\leftarrow
\operatorname{SelectParent}(\mathcal{A}_g,\mathcal{N})$
\STATE $\Pi_g\leftarrow(\mathcal{I},B,p_g,H_g,M_g,\sigma)$
\FOR{$k=1,\ldots,K$}
\STATE $y_{g,k}\sim q(\cdot\mid\Pi_g)$
\STATE $c_{g,k}\leftarrow \operatorname{Embed}_d(y_{g,k})$
\IF{$c_{g,k}=\bot \vee \operatorname{Admit}(c_{g,k})=0$}
\STATE $\mathcal{N}\leftarrow\mathcal{N}\cup\{(y_{g,k},c_{g,k})\}$
\ELSE
\STATE $\mathbf{m}_{g,k}\leftarrow\operatorname{Evaluate}(c_{g,k})$
\STATE $\mathcal{A}^{(r)}_{g}\leftarrow
\operatorname{Insert}(\mathcal{A}^{(r)}_{g},c_{g,k},\mathbf{m}_{g,k},\psi)$
\ENDIF
\ENDFOR
\STATE $\mathcal{A}_{g+1}\leftarrow
\operatorname{ParetoRefresh}(\operatorname{Migrate}(\mathcal{A}_{g}))$
\ENDFOR
\RETURN $\operatorname{Best}(\mathcal{A}_G)$
\end{algorithmic}
\end{algorithm}

\section{Experiments and Results}

\subsection{Experimental Setup}
\label{sec-experimental-setup}

\paragraph{Benchmarks and exposure.}
We evaluate CoEvoP\&R on eight ChiPBench Nangate45 designs~\cite{chipbench}, including
bp\_fe, bp\_be, swerv\_wrapper, ethernet, dft68, or1200, vga\_lcd, and
mor1kx. To test cross-technology generalization, we further use ICCAD 2015 Superblue
designs 1, 3, 4, 5, 7, 10, 16, and 18~\cite{iccad2015}, together with gcd,
ibex, and ariane implemented using the ASAP7 predictive 7\,nm standard-cell
library~\cite{asap7}. CoEvoP\&R-E uses feedback from the
same design on which it is evaluated. CoEvoP\&R-L follows
\textit{leave-one-design-out} (LODO) evaluation, where the target design is
excluded from objective evolution and prompt evidence. CoEvoP\&R-T zero-shot transfers the
best candidate from the ChiPBench evolution for cross-family test.

\paragraph{Baselines and evaluation flow.}
Baselines include analytical placement through DREAMPlace and DREAMPlace
4.0~\cite{dreamplace,dreamplace4}, expert HPWL, density,
RUDY~\cite{rudy}, and pin-density objective mixes, same-grammar BO-DSL
implemented with \textit{tree-structured Parzen estimator} (TPE)~\cite{tpe},
and shared-pipeline reruns of
AutoDMP~\cite{autodmp}, RoutePlacer~\cite{routeplacer},
EvoPlace~\cite{evoplace}, and LaMPlace~\cite{lamplace}. Comparable rows share
coordinate conversion, legalization, OpenROAD routing~\cite{openroad},
OpenTimer timing analysis~\cite{huang2015opentimer}, and metric extraction.

\paragraph{Evaluation and statistics.}
All results in Tables~\ref{tab:chipbench-main}
and~\ref{tab:generalization} report improvements over DREAMPlace, which
is normalized to zero. These results are averaged under reruns using three different seeds. The default LLM proposer is
\texttt{gpt-5.4}. All evolution and experiments run on one NVIDIA RTX A6000 GPU with 48\,GB
memory and one Intel Xeon Gold 5218R \textit{central processing unit} (CPU)
using 16 OpenROAD/OpenTimer worker processes. Section~\ref{sec:ablation} tests LLM model-family sensitivity.

\subsection{Main Results on ChiPBench Nangate45}
\label{sec:main-results}

Table~\ref{tab:chipbench-main} is the primary same-technology result, covering
CoEvoP\&R-E and CoEvoP\&R-L. CoEvoP\&R-E improves mean routed
wirelength and congestion by 16.9\% and 36.7\%, and improves WNS and TNS by
0.70 ns and 912 ns over DREAMPlace. The strongest non-CoEvo means reach
9.7\% rWL, 20.6\% congestion, 0.41 ns WNS, and 444 ns TNS, with the leading
method varying by metric. At the design level, LaMPlace remains strong on
bp\_fe and bp\_be wirelength, DREAMPlace 4.0 leads or1200 WNS, and CoEvoP\&R-E
supplies most of the best or second-best congestion and timing entries.
CoEvoP\&R-L keeps 13.2\% rWL, 31.4\% congestion, 0.51 ns WNS, and 637 ns TNS
under the LODO protocol. The best candidate in this experiment is zero-shot transferred to the cross-family evaluation in Section~\ref{sec:generalization}.

\begin{table}[!htbp]
\centering
\caption{Post-route transfer on Superblue and ASAP7. Entries report
three-seed mean improvements over DREAMPlace, which is normalized to zero. rWL and
Cong.\ are reductions in \textit{routed wirelength} and post-route
\textit{congestion}. WNS and TNS are post-route timing gains in ns.
\colorbox{rankfirst}{\textbf{Bold}} marks best means and
\colorbox{ranksecond}{\underline{underline}} marks second best.}
\label{tab:generalization}
\scriptsize
\renewcommand{\arraystretch}{0.82}
\setlength{\tabcolsep}{1.0pt}
\resizebox{\columnwidth}{!}{%
\begin{tabular}{llccccccccccc}
\toprule
\multicolumn{2}{c}{} & \multicolumn{8}{c}{Superblue} & \multicolumn{3}{c}{ASAP7} \\
\cmidrule(lr){3-10}\cmidrule(lr){11-13}
Method & Metric & sb1 & sb3 & sb4 & sb5 & sb7 & sb10 & sb16 & sb18 & gcd & ibex & ariane \\
\midrule
\multirow{4}{*}{DREAMPlace 4.0~\cite{dreamplace4}}
  & rWL (\%)         & -0.4 & -0.6 & +0.2 & -0.8 & -0.3 & +0.1 & -0.5 & +0.3 & +0.4 & -0.3 & -0.8 \\
  & Cong.\ (\%)      & +0.2 & -0.3 & -0.4 & +0.1 & -0.2 & -0.7 & -0.1 & +0.4 & +0.6 & +0.2 & -0.4 \\
  & WNS ($\Delta$ns) & \cellcolor{ranksecond}\underline{+4.4} & \cellcolor{rankfirst}\textbf{+16.8} & \cellcolor{ranksecond}\underline{+8.7} & \cellcolor{rankfirst}\textbf{+21.7} & \cellcolor{ranksecond}\underline{+5.2} & \cellcolor{ranksecond}\underline{+1.9} & \cellcolor{ranksecond}\underline{+4.7} & \cellcolor{rankfirst}\textbf{+8.3} & +0.06 & +0.18 & \cellcolor{ranksecond}\underline{+0.34} \\
  & TNS ($\Delta$ns) & \cellcolor{ranksecond}\underline{+16700} & +3370 & +5210 & +11300 & +9810 & \cellcolor{ranksecond}\underline{+7040} & \cellcolor{rankfirst}\textbf{+31400} & +4320 & +32 & +187 & \cellcolor{rankfirst}\textbf{+940} \\
\midrule
\multirow{4}{*}{RoutePlacer~\cite{routeplacer}}
  & rWL (\%)         & +0.1 & -0.3 & +0.1 & +0.3 & -0.1 & +0.9 & -0.1 & +0.2 & +0.4 & -0.6 & +1.8 \\
  & Cong.\ (\%)      & +3.1 & +1.7 & +4.6 & +3.9 & +3.9 & \cellcolor{rankfirst}\textbf{+43.2} & +1.4 & +3.7 & \cellcolor{ranksecond}\underline{+6.4} & \cellcolor{rankfirst}\textbf{+24.6} & \cellcolor{rankfirst}\textbf{+28.4} \\
  & WNS ($\Delta$ns) & -0.6 & -0.4 & +0.2 & -0.4 & -0.8 & -1.2 & -0.3 & -0.7 & -0.02 & -0.14 & -0.34 \\
  & TNS ($\Delta$ns) & -1400 & -800 & +200 & -1200 & -1600 & -3400 & -800 & -1400 & -12 & -87 & -294 \\
\midrule
\multirow{4}{*}{EvoPlace~\cite{evoplace}}
  & rWL (\%)         & -2.4 & -3.6 & +0.4 & -2.8 & +1.4 & -3.2 & -1.8 & +0.6 & +1.2 & -8.4 & -13.4 \\
  & Cong.\ (\%)      & -3.4 & -8.6 & -2.4 & -4.2 & +0.4 & -12.4 & -6.8 & -1.2 & +0.4 & -14.2 & -18.4 \\
  & WNS ($\Delta$ns) & +0.2 & +0.4 & +0.4 & -0.4 & +0.6 & -0.8 & -0.4 & +0.4 & +0.02 & -0.24 & -0.48 \\
  & TNS ($\Delta$ns) & -600 & -1400 & +400 & -1200 & +300 & -2200 & -1400 & +200 & +4 & -124 & -487 \\
\midrule
\multirow{4}{*}{LaMPlace~\cite{lamplace}}
  & rWL (\%)         & -4.6 & -6.8 & \cellcolor{rankfirst}\textbf{+14.2} & -8.4 & +0.4 & -2.4 & \cellcolor{rankfirst}\textbf{+11.8} & \cellcolor{rankfirst}\textbf{+22.4} & -6.8 & -18.4 & -22.4 \\
  & Cong.\ (\%)      & \cellcolor{ranksecond}\underline{+25.2} & +19.9 & +4.9 & +19.8 & \cellcolor{ranksecond}\underline{+23.0} & +24.8 & +7.7 & \cellcolor{ranksecond}\underline{+28.2} & +2.4 & -8.7 & -14.2 \\
  & WNS ($\Delta$ns) & +2.4 & +6.8 & +3.4 & \cellcolor{ranksecond}\underline{+13.4} & +3.6 & +1.2 & +2.8 & +0.4 & +0.04 & -0.16 & -0.32 \\
  & TNS ($\Delta$ns) & +2400 & +6800 & +2400 & +8400 & +3800 & +1400 & +1800 & +1200 & -18 & -94 & -246 \\
\midrule
\multirow{4}{*}{CoEvoP\&R-E}
  & rWL (\%)         & \cellcolor{rankfirst}\textbf{+5.4} & \cellcolor{rankfirst}\textbf{+6.2} & \cellcolor{ranksecond}\underline{+4.8} & \cellcolor{rankfirst}\textbf{+7.2} & \cellcolor{rankfirst}\textbf{+5.4} & \cellcolor{rankfirst}\textbf{+3.6} & \cellcolor{ranksecond}\underline{+6.4} & \cellcolor{ranksecond}\underline{+4.2} & \cellcolor{rankfirst}\textbf{+5.4} & \cellcolor{rankfirst}\textbf{+8.2} & \cellcolor{rankfirst}\textbf{+10.6} \\
  & Cong.\ (\%)      & \cellcolor{rankfirst}\textbf{+26.4} & \cellcolor{rankfirst}\textbf{+21.4} & \cellcolor{rankfirst}\textbf{+15.8} & \cellcolor{rankfirst}\textbf{+22.4} & \cellcolor{rankfirst}\textbf{+24.4} & \cellcolor{ranksecond}\underline{+34.7} & \cellcolor{rankfirst}\textbf{+9.8} & \cellcolor{rankfirst}\textbf{+30.4} & \cellcolor{rankfirst}\textbf{+14.6} & \cellcolor{ranksecond}\underline{+22.4} & \cellcolor{ranksecond}\underline{+26.8} \\
  & WNS ($\Delta$ns) & \cellcolor{rankfirst}\textbf{+5.8} & \cellcolor{ranksecond}\underline{+14.8} & \cellcolor{rankfirst}\textbf{+9.4} & +12.8 & \cellcolor{rankfirst}\textbf{+7.6} & \cellcolor{rankfirst}\textbf{+2.6} & \cellcolor{rankfirst}\textbf{+6.2} & \cellcolor{ranksecond}\underline{+6.4} & \cellcolor{rankfirst}\textbf{+0.24} & \cellcolor{rankfirst}\textbf{+0.42} & \cellcolor{rankfirst}\textbf{+0.68} \\
  & TNS ($\Delta$ns) & \cellcolor{rankfirst}\textbf{+18800} & \cellcolor{rankfirst}\textbf{+38400} & \cellcolor{rankfirst}\textbf{+18700} & \cellcolor{rankfirst}\textbf{+67200} & \cellcolor{rankfirst}\textbf{+36400} & \cellcolor{rankfirst}\textbf{+8400} & \cellcolor{ranksecond}\underline{+14600} & \cellcolor{rankfirst}\textbf{+6800} & \cellcolor{rankfirst}\textbf{+148} & \cellcolor{rankfirst}\textbf{+468} & \cellcolor{ranksecond}\underline{+824} \\
\midrule
\multirow{4}{*}{CoEvoP\&R-T}
  & rWL (\%)         & \cellcolor{ranksecond}\underline{+1.4} & \cellcolor{ranksecond}\underline{+3.8} & +2.6 & \cellcolor{ranksecond}\underline{+4.2} & \cellcolor{ranksecond}\underline{+3.2} & \cellcolor{ranksecond}\underline{+1.8} & +3.6 & +2.4 & \cellcolor{ranksecond}\underline{+2.4} & \cellcolor{ranksecond}\underline{+3.8} & \cellcolor{ranksecond}\underline{+5.2} \\
  & Cong.\ (\%)      & +20.4 & \cellcolor{ranksecond}\underline{+20.4} & \cellcolor{ranksecond}\underline{+10.4} & \cellcolor{ranksecond}\underline{+20.4} & +18.4 & +25.4 & \cellcolor{ranksecond}\underline{+8.4} & +24.6 & \cellcolor{ranksecond}\underline{+8.4} & +19.4 & +25.4 \\
  & WNS ($\Delta$ns) & \cellcolor{ranksecond}\underline{+4.6} & +8.2 & \cellcolor{ranksecond}\underline{+8.8} & +7.4 & \cellcolor{ranksecond}\underline{+5.4} & \cellcolor{ranksecond}\underline{+2.2} & \cellcolor{ranksecond}\underline{+5.4} & +4.2 & \cellcolor{ranksecond}\underline{+0.08} & \cellcolor{ranksecond}\underline{+0.22} & \cellcolor{ranksecond}\underline{+0.34} \\
  & TNS ($\Delta$ns) & +2400 & \cellcolor{ranksecond}\underline{+18400} & \cellcolor{ranksecond}\underline{+9600} & \cellcolor{ranksecond}\underline{+32400} & \cellcolor{ranksecond}\underline{+14200} & \cellcolor{ranksecond}\underline{+7400} & +7600 & \cellcolor{ranksecond}\underline{+4600} & \cellcolor{ranksecond}\underline{+58} & \cellcolor{ranksecond}\underline{+204} & +620 \\
\bottomrule
\end{tabular}
}
\end{table}

\subsection{Generalization}
\label{sec:generalization}

Table~\ref{tab:generalization} reports Superblue and ASAP7 transfer results
for CoEvoP\&R-E and CoEvoP\&R-T.

On Superblue, E averages 5.4\% rWL, 23.2\% congestion reduction, 8.2 ns WNS,
and 26,163 ns TNS gain. T retains 2.9\%, 18.6\%, 5.8 ns, and 12,075 ns, or
46\% to 80\% of E. On ASAP7, LaMPlace regresses while both CoEvoP\&R variants
remain positive, supporting symbolic cross-node transfer. Superblue1 is the
clearest transfer-limit case.

\begin{figure}[h]
\centering
\includegraphics[width=\columnwidth]{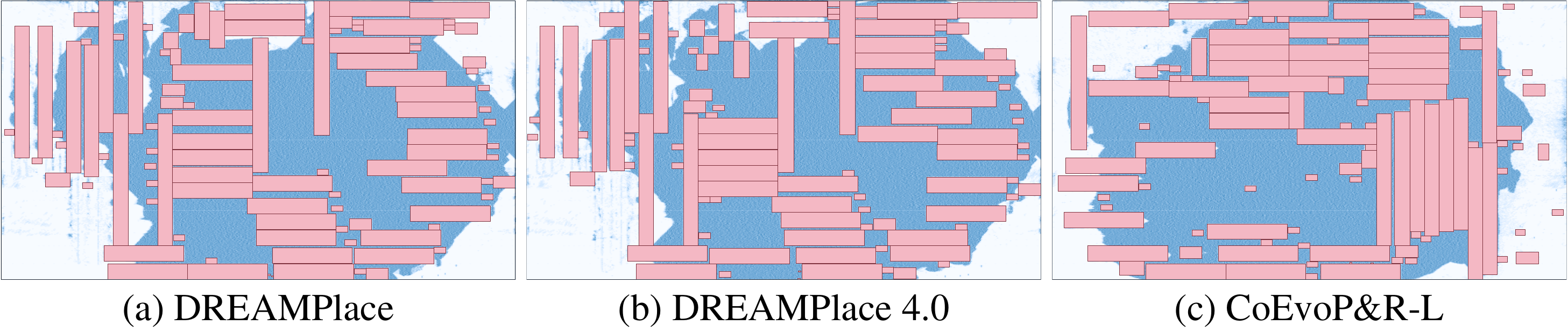}\\[0.1em]
{\scriptsize
\setlength{\tabcolsep}{1pt}
\renewcommand{\arraystretch}{0.95}
\begin{tabular}{@{}ccc@{}}
\parbox{0.33\columnwidth}{\centering rWL=0.0, Cong.=0.0\\WNS=0.0, TNS=0}
& \parbox{0.33\columnwidth}{\centering rWL=$-0.5$, Cong.=$-0.1$\\WNS=+4.7, TNS=+31{,}400}
& \parbox{0.33\columnwidth}{\centering rWL=+3.6, Cong.=+8.4\\WNS=+5.4, TNS=+12{,}800}
\end{tabular}}
\caption{Superblue16 qualitative placement comparison. Metrics below each panel are improvements over the DREAMPlace reference.}
\Description{Three-panel Superblue16 placement comparison showing DREAMPlace,
DREAMPlace 4.0 timing-driven, and CoEvoP\&R-L with matching placement and
density overlays plus rWL, Cong., WNS, and TNS values.}
\label{fig:placement-qualitative}
\end{figure}

Figure~\ref{fig:placement-qualitative} visualizes Superblue16 under matched
placement rendering. CoEvoP\&R-L improves rWL/Cong.\ by 3.6\%/8.4\% and gains
5.4 ns WNS and 12,800 ns TNS over DREAMPlace, while DREAMPlace 4.0's
timing-driven flow leads on TNS (+31,400 ns) at the cost of slightly worse
rWL and congestion.

\subsection{Timing proxy Audit}

\begin{figure}[h]
\centering
\includegraphics[width=\columnwidth]{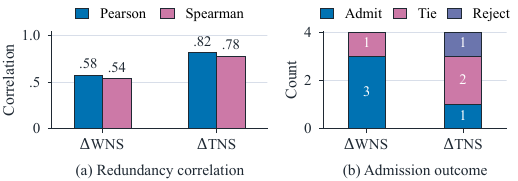}
\caption{Timing-proxy admission audit. Grouped bars show median Pearson and
Spearman correlation with HPWL movement.}
\Description{Two-panel bar chart for timing proxy audit. Delta WNS has Pearson
0.58 and Spearman 0.54, with three Admit outcomes and one Tie outcome. Delta
TNS has Pearson 0.82 and Spearman 0.78, with one Admit, two Tie, and one Reject
outcome.}
\label{fig:proxy-audit}
\end{figure}

Figure~\ref{fig:proxy-audit} audits bp\_fe, swerv, ethernet, and or1200
under 100 to 2000 \textit{database unit} (DBU) perturbations. The left panel measures HPWL redundancy,
and the right panel counts admission outcomes. Admit uses the proxy in
candidate selection, Tie uses it only to break close cases, and Reject excludes
it for that metric. WNS has lower HPWL redundancy and is admitted on three of
four designs, making it the primary timing proxy. TNS tracks HPWL more strongly
and acts mainly as a tie-breaker. The ungated timing-proxy row in
Table~\ref{tab:ablation-efficiency} tests this admission rule.

\subsection{Ablation Studies and LLM Sensitivity}
\label{sec:ablation}

Table~\ref{tab:ablation-efficiency} isolates the main sources of performance
in CoEvoP\&R. Random DSL evolution and one-shot LLM sampling lose most of the
routing and timing gains, showing that archive-conditioned iteration is
essential. Tier-A-only selection and direct Tier~C evaluation both underperform
Full CoEvoP\&R, indicating that cost-scaled evidence allocation is more useful
than either cheap-only or route-all evaluation. Scalar-history and metrics-only
archives also degrade, showing that component traces and failure memory provide
actionable search context. The model variants preserve the main advantage,
with stronger LLMs improving validity and score under the same prompts,
mutation schedules, and budgets.

\begin{table}[H]
\caption{Ablation and LLM sensitivity on the ChiPBench stress panel
bp\_fe, swerv\_wrapper, ethernet, and or1200. R/C and W/T denote routed
wirelength/congestion reductions and WNS/TNS gains. Valid/Acc.\ denote valid
and admitted candidates. Evals and GPU-h denote evaluations and GPU-hours.
Darker blue indicates larger degradation from Full CoEvoP\&R.}
\label{tab:ablation-efficiency}
\centering
\scriptsize
\setlength{\tabcolsep}{1.5pt}
\renewcommand{\arraystretch}{0.88}
\resizebox{\columnwidth}{!}{%
\begin{tabular}{p{0.34\columnwidth}cccccc}
\toprule
Configuration & Valid/Acc. & Score & R/C & W/T & Tier-C & Cost \\
& & & (\%) & ($\Delta$ns) & evals & (GPU-h) \\
\midrule
\rowcolor{degfull}
\cellcolor{white}\textbf{Full CoEvoP\&R} & 68/12 & 4.87 & 20.5/45.0 & +0.64/+890 & 24 & 32 \\
\midrule
\multicolumn{7}{l}{\textit{Proposal and objective search}} \\
\rowcolor{degworst}
\cellcolor{white}Random DSL evolution       & \textcolor{white}{42/4}  & \textcolor{white}{1.24} & \textcolor{white}{2.6/4.8}   & \textcolor{white}{+0.08/+60}  & \textcolor{white}{24} & \textcolor{white}{26} \\
\rowcolor{degbad}
\cellcolor{white}One-shot LLM sampling      & \textcolor{white}{58/8}  & \textcolor{white}{2.14} & \textcolor{white}{7.2/13.4}  & \textcolor{white}{+0.22/+264} & \textcolor{white}{24} & \textcolor{white}{16} \\
\rowcolor{degfair}
\cellcolor{white}Score-only LLM evolution   & 62/10 & 3.42 & 14.2/26.4 & +0.40/+454 & 24 & 29 \\
\midrule
\multicolumn{7}{l}{\textit{Evaluation and evidence policy}} \\
\rowcolor{degbad}
\cellcolor{white}Prompt-only free-form interface & \textcolor{white}{24/4} & \textcolor{white}{1.86} & \textcolor{white}{5.8/12.4}  & \textcolor{white}{+0.15/+164} & \textcolor{white}{24} & \textcolor{white}{27} \\
\rowcolor{degpoor}
\cellcolor{white}Tier-A-only selection      & 68/18 & 2.68 & 9.4/15.4  & +0.22/+268 & 0  & 5 \\
\rowcolor{degfair}
\cellcolor{white}Direct Tier C (all valid)  & 68/8  & 3.14 & 12.0/23.6 & +0.32/+498 & 68 & 64 \\
\rowcolor{deggood}
\cellcolor{white}Ungated timing proxy       & 68/14 & 3.86 & 16.4/34.6 & +0.46/+588 & 24 & 33 \\
\midrule
\multicolumn{7}{l}{\textit{Archive state}} \\
\rowcolor{degpoor}
\cellcolor{white}Champion-only scalar history & 62/10 & 2.42 & 9.6/17.6  & +0.26/+312 & 24 & 31 \\
\rowcolor{degfair}
\cellcolor{white}Top-$k$ scalar archive       & 64/11 & 3.14 & 13.6/26.4 & +0.36/+448 & 24 & 32 \\
\rowcolor{deggood}
\cellcolor{white}MAP-Elites, metrics only     & 66/12 & 3.68 & 17.0/34.4 & +0.46/+588 & 24 & 32 \\
\midrule
\multicolumn{7}{l}{\textit{Model family sensitivity}} \\
\rowcolor{degfull}
\cellcolor{white}\texttt{gpt-5.4} (default)    & 68/12 & 4.87 & 20.5/45.0 & +0.64/+890 & 24 & 32 \\
\rowcolor{deggood}
\cellcolor{white}\texttt{gpt-4.1-mini}         & 62/10 & 3.87 & 16.4/36.0 & +0.49/+668 & 24 & 26 \\
\rowcolor{degvgood}
\cellcolor{white}\texttt{claude-opus-4-8}      & 66/12 & 4.62 & 19.4/43.2 & +0.60/+836 & 24 & 35 \\
\rowcolor{degfair}
\cellcolor{white}\texttt{qwen3.7-max}          & 58/9  & 3.24 & 13.8/28.4 & +0.42/+528 & 24 & 28 \\
\bottomrule
\end{tabular}
}
\end{table}

\section{Conclusion}
\label{sec:conclusion}

CoEvoP\&R evolves executable placement objectives with bounded analytical
placement and routed feedback, turning routing and timing evidence into
readable differentiable objectives for DREAMPlace. Across ChiPBench, Superblue,
and ASAP7, the evolved objectives improve routed wirelength, congestion, and
timing. On ChiPBench, target evolution reaches 16.9\% rWL reduction, 36.7\%
congestion reduction, 0.70 ns WNS gain, and 912 ns TNS gain over DREAMPlace. The
results support LLM-guided objective evolution as an inspectable interface
between differentiable placement and downstream evaluation. A natural next step
is to extend the archive and validation interface toward multi-corner timing,
power, and signoff-accurate routing feedback.

\bibliographystyle{ACM-Reference-Format}
\bibliography{sample-base}

\end{document}